\RequirePackage{amsmath}
\documentclass[runningheads]{llncs}

\usepackage[utf8]{inputenc}
\usepackage[english]{babel}
 
\usepackage{xspace}
\usepackage{graphicx}
\newcommand{\mycaptionof}[2]{\captionof{#1}{#2}}
\usepackage[utf8]{inputenc}
\newcommand{\btheta}{{\mathbb \theta}}
\DeclareUnicodeCharacter{1EF3}{\`y}
\newcommand{\lbs}{\ensuremath{B}}      %

\newcommand{\lepochs}{\ensuremath{E}}  %
\newcommand{\clientfrac}{\ensuremath{C}}    %

\newcommand{\loss}{\ell}
\newcommand{\SUB}[1]{\ENSURE \hspace{-0.15in} \textbf{#1}}
\newcommand{\algfont}[1]{\texttt{#1}}

\newcommand{\nc}{K}
\newcommand{\pp}{\mathcal{P}}

\newcommand{\R}{\ensuremath{\mathbb{R}}}

\newcommand{\grad}{\triangledown}

\newcommand{\bW}{\mathbf{W}}

\newcommand{\bH}{\mathbf{H}}

\newcommand{\bA}{\mathbf{A}}

\renewcommand{\[}{\begin{equation*}}
\renewcommand{\]}{\end{equation*}}

\usepackage{algorithm}
\usepackage[noend]{algorithmic}
\usepackage{lipsum}
\usepackage{booktabs}
\usepackage[utf8]{inputenc} 
\usepackage[T1]{fontenc}    
\usepackage{hyperref}       
\usepackage{url}            
\usepackage{booktabs}       
\usepackage{amsfonts}       
\usepackage{amsmath}
\usepackage{nicefrac}       
\usepackage{microtype}      
\usepackage{graphicx}
\usepackage{subfig}
\usepackage{amssymb}

\usepackage{tabularx}
\usepackage{graphicx}

\begin{document}
\title{No Peek: A Survey of private distributed deep learning}

%
%
%
\author{Praneeth Vepakomma\thanks{Corresponding author e-mail: vepakom@mit.edu } \and
Tristan Swedish \and
Ramesh Raskar  \and
 Otkrist Gupta\and
 Abhimanyu Dubey}\institute{Massachusetts Institute of Technology\\ Cambridge, MA 02139, U.S.A}
\authorrunning{No Peek: A Survey of private distributed deep learning}
%
%
\maketitle              

\vspace{-0.6cm}
\begin{abstract}
We survey distributed deep learning models for training or inference without accessing raw data from clients. These methods aim to protect confidential patterns in data while still allowing servers to train models. The distributed deep learning methods of federated learning, split learning and large batch stochastic gradient descent are compared in addition to private and secure approaches of differential privacy, homomorphic encryption, oblivious transfer and garbled circuits in the context of neural networks. We study their benefits, limitations and trade-offs with regards to computational resources, data leakage and communication efficiency and also share our anticipated future trends.

\end{abstract}
\section{Introduction}
Emerging technologies in domains such as biomedicine, health and finance benefit from distributed deep learning methods which can allow multiple entities to train a deep neural network without requiring data sharing or resource aggregation at one single place. In particular, we are interested in distributed deep learning approaches that bridge the gap between distributed data sources (clients) and a powerful centralized computing resource (server) under the constraint that local data sources of clients are not allowed to be shared with the server or amongst other clients. 

We survey and compare such distributed deep learning techniques and classify them across various dimensions of level and type of protection offered, model performance and resources required such as memory, time, communications bandwidth and synchronization requirements. We introduce the terminology of ‘no peek’ to refer to distributed deep learning techniques that do not share their data in raw form. We note that such no peek techniques allow the server to train models without 'peeking at', or directly observing, raw data belonging to clients.

Additionally, we survey some generic approaches to protecting data and models. Some of these approaches have already been used in combination with distributed deep learning methods that possess varying levels of the no peek property. These generic approaches include de-identification methods like anonymization \cite{li2007t}, obfuscation methods like differential privacy \cite{dwork2014algorithmic,Dwork2006CalibratingNT,Dwork:2011:FFP:1866739.1866758} and cryptographic techniques like homomorphic encryption \cite{aono2018privacy,Wu2012UsingHE,Lu2016UsingFH} and secure multi-party computation (MPC) protocols like oblivious transfer \cite{orlandi2007oblivious,liu2017oblivious} and garbled circuits \cite{rosulek2017improvements}. 

In the rest of the paper, we focus on distributed deep learning techniques such as splitNN \cite{gupta2018distributed,vepakomma2018split}, large batch synchronous stochastic gradient descent (SGD) \cite{chen2016revisiting,konevcny2016federated}, federated learning\cite{mcmahan2016communication} and other variants \cite{dean2012large,wen2017terngrad,das2016distributed,ooi2015singa} in the context of protecting data and models.

\subsection{No peek rule} We refer to techniques of distributed deep learning that do not look at raw data once it leaves the client as satisfying the property of 'no peek'. No peek is necessitated by trust and regulatory issues. For example, hospitals are typically not allowed to share data with for-profit entities due to trust issues. They also are unable to share it with external entities (data cannot physically leave the premises) due to limited consent of the patients, and regulations such as HIPAA \cite{annas2003hipaa,centers2003hipaa,mercuri2004hipaa,gostin2009beyond,luxton2012mhealth} that prevent sharing many aspects of the data to external entities. Some techniques go a step ahead by also not revealing details of the model architecture as well. In these techniques, neither the server nor client can access the details of the other's architecture or weights.  

\subsection{What needs to be protected}
Protection mechanisms in the context of distributed deep learning should protect various aspects of datasets such as 

\begin{enumerate}
\item Input features
\item Output labels or responses 
\item Model details including the architecture, parameters and loss function
\item Identifiable information such as which party contributed to a specific record
\end{enumerate}

\subsection{Computational Goals} It is also quite important that any mechanism that aims to protect these details also preserves utility of the model above an acceptable level. These goals are to be ideally achieved at a low cost with regards to 

\begin{enumerate}
\item Memory 
\item Computational time 
\item Communications bandwidth
\item Synchronization

\end{enumerate}
As shown in Fig 1. below smaller hospitals or tele-healthcare screening centers do not acquire an enormous number of diagnostic images and they could also be limited by diagnostic manpower. A distributed machine learning method for diagnosis in this setting should ideally not share any raw data (no peek) and at same time achieve high accuracy while using significantly lower resources. This helps smaller hospitals to effectively serve those in need while benefiting from decentralized entities. 
\begin{table*}[htbp]
\centering
\begin{tabular}{|p{2.9cm}|p{2.9cm}|p{2.0cm}|p{2.3cm}|p{3.1cm}|}
\hline
\textbf{Distributed Method}                                                    & \textbf{Partial/Full Leakage}                                                                                          & \textbf{\begin{tabular}[c]{@{}l@{}}Differential \\ Privacy\end{tabular}}                                               & \textbf{\begin{tabular}[c]{@{}l@{}}Homomorphic\\ Encryption\end{tabular}}     & \textbf{\begin{tabular}[c]{@{}l@{}}Oblivious Transfer,\\ Garbled Circuits\end{tabular}}                                   \\ \hline
\textbf{Distributed NN}                                                        & \begin{tabular}[c]{@{}l@{}}{[}Dean2012, Wen2017,\\Das2016, Ooi2015{]},\\ Ben2018{]} \end{tabular} & \begin{tabular}[c]{@{}l@{}}{[}Hynes2018, \\Abadi2016,\\ Shokri2015, \\ Papernot2016{]}\end{tabular} & \begin{tabular}[c]{@{}l@{}}{[}Juvekar2018, \\Gilad2016{]}\end{tabular} & \begin{tabular}[c]{@{}l@{}}{[}Rouhani2017, \\Mohassel2017,\\Riazi2018, \\ Orlandi2007{]}\end{tabular} \\ \hline

\textbf{\begin{tabular}[c]{@{}l@{}}Large Batch\\ Synchronous SGD\end{tabular}} & \begin{tabular}[c]{@{}l@{}}{[}Konečný2015, \\Chen2016{]}\end{tabular}                                           &                                                                                                                        &                                                                               &                                                                                                                           \\ \hline
\textbf{Federated Learning}                                                    & \begin{tabular}[c]{@{}l@{}}{[}McMahan2017, \\Nock2018{]}\end{tabular}                                           & {[}Geyer2017{]}                                                                                                        & \begin{tabular}[c]{@{}l@{}}{[}Aono2018,\\ Hardy2017{]}\end{tabular}     & {[}Bonawitz2016{]}                                                                                                        \\ \hline
\textbf{SplitNN}                                                               & {[}Gupta2018, Vepakomma2018{]}                                                                                                        &                                                                                                                        &                                                                               &                                                                                                                           \\ \hline
\end{tabular}
 \centering
 \caption{This is a survey of distributed deep learning methods with decreasing levels of leakage from distributed NN to splitNN. Hybrid approaches of these techniques and differential privacy, homomorphic encryption and MPC are also included. The citations for these 9 groups have been grouped separately with subtitles in the references section for convenience.}
\label{resource_table}
\end{table*}

\section{No peek approaches for distributed deep learning}In table 1 we provide corresponding references to various combinations of distributed deep learning techniques along with generic approach\begin{figure}[!tbp]
  \centering
  \begin{minipage}[b]{0.45\textwidth}
    \includegraphics[width=\textwidth]{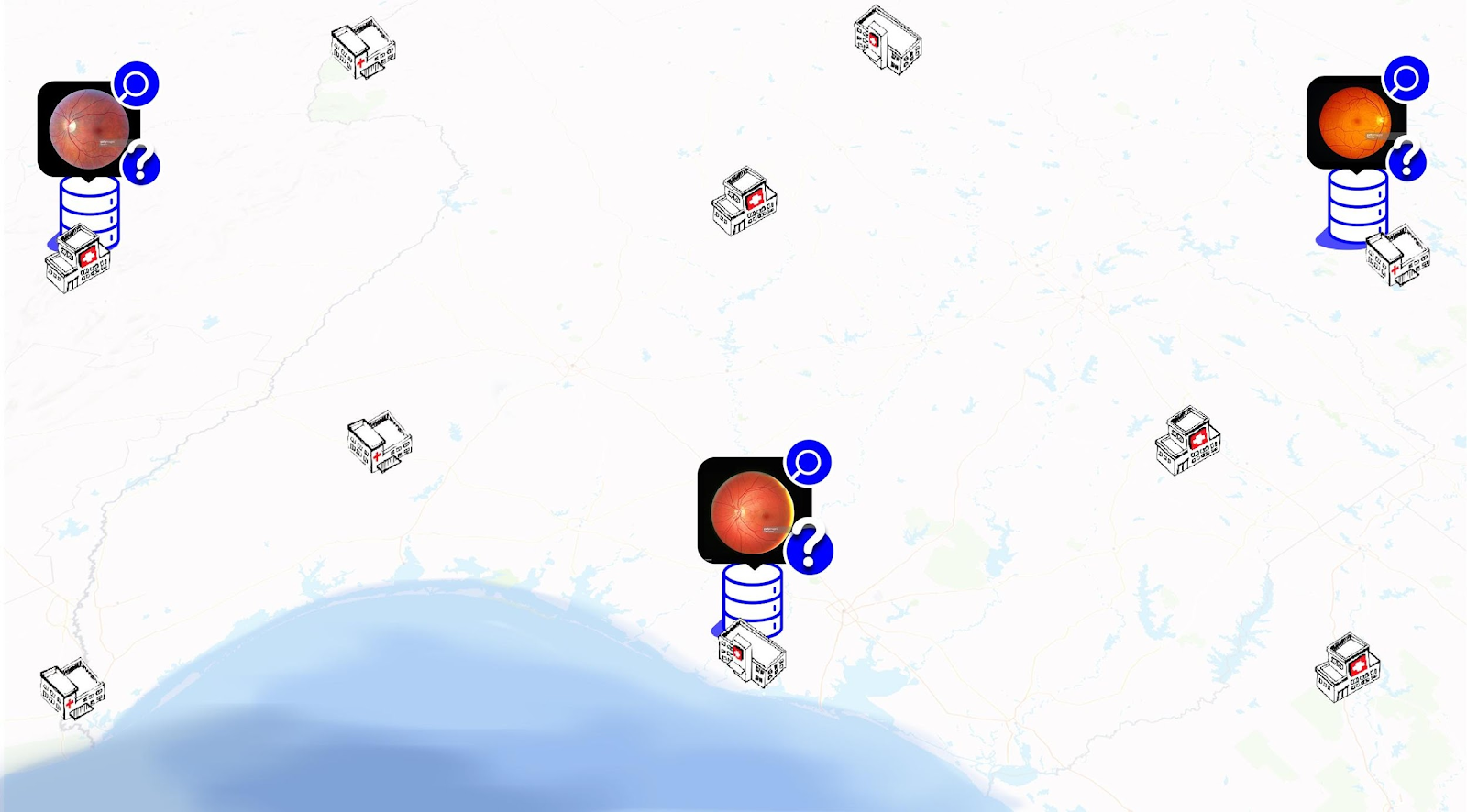}
    \caption{Non-cooperating health units}
  \end{minipage}
  \hfill
  \begin{minipage}[b]{0.45\textwidth}
    \includegraphics[width=\textwidth]{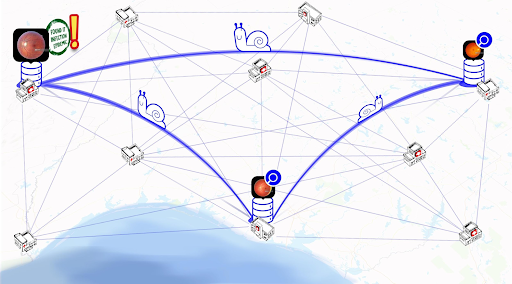}
    \caption{Distributed learning without raw data sharing}
  \end{minipage}
\end{figure}es for protection that are not specific to deep learning such as differential privacy homomorphic encryption and secure multi-party computation. The distributed deep learning techniques such as splitNN, federated learning and large batch synchronous SGD are `no peek'. In addition splitNN also protects model details of the architecture and weights, unlike the other techniques. We detail this below in table 2 in terms of the levels of protection offered on data, intermediate representations and hyperparameters that include the deep learning architecture and learnt weights.

In federated learning and large batch synchronous SGD, the architecture and parameters are shared between the client and server along with intermediate representations of the model that include the gradients, activations and weight updates which are shared during the learning process. Although the data is not explicitly shared in raw form in these two techniques, works like \cite{hitaj2017deep} have shown that raw data can be reconstructed up to an approximation by adversaries, especially given the fact that the architecture and parameters are not completely protected.\vspace{1em} SplitNN \cite{gupta2018distributed} has an added advantage in this context in that it does not share the architecture and weights of the model. The protection offered by splitNN lies in the compact representations found in deeper layers of neural networks and the difficulty of recovering the underlying data from these representations without knowing the model weights used to produce them. Such representations form after passing the data through numerous activations whose inverse in the case of ReLU are nonlinear and ill-defined (the inverse of a zero-valued ReLU can map to any negative number). Such representations have been shown to preserve information important for certain tasks (path following \cite{Swedish_2018_CVPR_Workshops}), without revealing information about the underlying data (such as image features in a 3D coordinate system). The intermediate representation shared by splitNN also requires minimal bandwidth in comparison to federated learning and large batch synchronous SGD, as only the activations from one layer of the client called the cut layer are shared with the server without any associated functions required to invert them back to raw data.
\vspace{-1em}
 \begin{table}[]
 \centering
\begin{tabular}{|l|l|l|l|}

\hline
\textbf{No Peek Deep Learning}                                                        & \textbf{Data revealed} & \textbf{\begin{tabular}[c]{@{}l@{}}Hyperparameters\\ revealed\end{tabular}} & \textbf{\begin{tabular}[c]{@{}l@{}}Intermediate \\ representation\\ revealed\end{tabular}} \\ \hline
\textbf{\begin{tabular}[c]{@{}l@{}}Large Batch\\ Synchronous SGD\end{tabular}} & No            & Yes                                                                                           & Yes                                                                              \\ \hline \textbf{Federated Learning}                                                    & No            & Yes                                                                                           & Yes                                                                              \\ \hline
\textbf{SplitNN}                                                               & No            & No                                                                                            & Yes                                                                              \\ \hline
\end{tabular}

\caption{In this table, we compare the level of privacy offered over data, model architecture, model parameters and intermediate representations by techniques like federated learning, large batch synchronous SGD and splitNN. On all these aspects, splitNN out performs federated learning or large batch synchronous SGD.}
\end{table}
\vspace{-3em}
In table 3, we compare these techniques based on resources required such as computations, communication bandwidth, memory and synchronization. We categorize the techniques across these dimensions as having low, medium and high requirements. As shown, splitNN requires the lowest resources on the client side. This is because the architecture is cut (arbitrary shape and not necessarily vertical) at a layer where the computations are only performed up to that cut on the client side. The rest of the computations happen on the server side. The experimental results in \cite{gupta2018distributed}, quantify these comparisons.
\section{Federated Learning}\vspace{-0.40cm} \textbf{Key idea: }In this approach the clients download the current model from central server and improve it by updating their model weights based on their local data. The client model parameter updates are aggregated to generate server model. This model is again downloaded by the clients and the process continues. There is no explicit sharing of raw data in this setup.
\begin{algorithm}[H]
\begin{algorithmic}
\SUB{Server executes at round $t\geq 0$:}
   \STATE Distribute $\bW_{t}$ to a subset $S_t$ of $n_t$ clients
     \FOR{each client $k \in S_t$ \textbf{in parallel}}
       \STATE $\bH^k_{t} \leftarrow \text{ClientUpdate}(k, w_t)$ 
     \ENDFOR
     \STATE Set $\bH_t := \tfrac{1}{n_t}\textstyle\sum_{i \in S_t}  \bH^i_{t}$
     \STATE Set $\bW_{t+1} = \bW_{t} + \eta_t \bH_{t}$
   \STATE

 \SUB{ClientUpdate($k, \bW_{t}$):}\ \ \  // \emph{Run on client $k$}
  \STATE $\mathcal{B} \leftarrow$ (split $\pp_k$ into batches of size $\lbs$)
  \STATE Set $\bW^k_{t} = \bW_{t}$
  \FOR{each local epoch $i$ from $1$ to $\lepochs$}
    \FOR{batch $b \in \mathcal{B}$}
      \STATE $\bW^k_{t} \leftarrow \bW^k_{t} - \eta \grad \loss(\bW^k_{t}; b)$
    \ENDFOR
 \ENDFOR
 \STATE return $\bH^k_t = \bW^k_{t} - \bW_{t}$ to server
\end{algorithmic}
\mycaptionof{algorithm}{\algfont{Naive Federated Learning}\xspace. Goal: To learn $\bW\in \R^{d_1\times d_2}$ from data stored across a large number of clients. The $\nc$
  clients are indexed by $k$; $\lbs$ is the local minibatch size,
  $\lepochs$ is the number of local epochs, and $\eta$ is the learning
  rate.}
\end{algorithm}
\vspace{-3em}

\begin{algorithm}[H]
\begin{algorithmic}
\SUB{Server executes:}
   \STATE initialize $w_0$
   \FOR{each round $t = 1, 2, \dots$}
     \STATE $m \leftarrow \max(\clientfrac\cdot K, 1)$
     \STATE $S_t \leftarrow$ (random set of $m$ clients)
     \FOR{each client $k \in S_t$ \textbf{in parallel}}
       \STATE $w_{t+1}^k \leftarrow \text{ClientUpdate}(k, w_t)$ 
     \ENDFOR
     \STATE $w_{t+1} \leftarrow \sum_{k=1}^\nc \frac{n_k}{n} w_{t+1}^k$
   \ENDFOR
   \STATE

 \SUB{ClientUpdate($k, w$):}\ \ \  // \emph{Run on client $k$}
  \STATE $\mathcal{B} \leftarrow$ (split $\pp_k$ into batches of size $\lbs$)
  \FOR{each local epoch $i$ from $1$ to $\lepochs$}
    \FOR{batch $b \in \mathcal{B}$}
      \STATE $w \leftarrow w - \eta \grad \loss(w; b)$
    \ENDFOR
 \ENDFOR
 \STATE return $w$ to server
\end{algorithmic}
\mycaptionof{algorithm}{(Communication-Efficient Learning of Deep Networks from Decentralized Data):\algfont{Federated Averaging}\xspace. The $\nc$
  clients are indexed by $k$; $\lbs$ is the local minibatch size,
  $\lepochs$ is the number of local epochs, and $\eta$ is the learning
  rate.}
\end{algorithm}
\vspace{-2em}
\subsection{Benefits} There is no explicit sharing of raw data. It has been shown in the convex case with IID data that in the worst-case, the global model produced is no better than training a model on a single client \cite{konevcny2016federated,mcmahan2016communication}.
\subsection{Limitations} Performance drops sharply when local data with clients is non i.i.d. That said, recent work in \cite{zhao2018federated} on federated learning in this setting shows positive results. It also requires large network bandwidth, memory and computation requirements on the client side depending on size of model, computation needs of complete forward and backward propagation. Advanced compression methods can be used instead to reduce this overload. There has been active recent work for neural network compression such as \cite{lin2017deep,han2015deep,louizos2017bayesian}.  These works can thereby reduce the costs for communication bandwidth when used in distributed learning. Federated learning has no theoretical guarantees or trade-offs of privacy or security to date.
 
\begin{table}[]
\centering
\begin{tabular}{|l|l|l|l|l|}
\hline
\textbf{\begin{tabular}[c]{@{}l@{}}Client Resources\\ Required\end{tabular}}    & \textbf{Compute} & \textbf{Bandwidth} & \textbf{Memory} & \textbf{Synchronization}                                                                                                                    \\ \hline
\textbf{\begin{tabular}[c]{@{}l@{}}Large Batch \\ Synchronous SGD\end{tabular}} & High             & High               & High            & \begin{tabular}[c]{@{}l@{}}Synchronous updates with backup\\ workers to compensate slow machines.\end{tabular} \\ \hline
\textbf{Federated Learning}                                                     & Medium           & Medium             & High            & Synchronous client-server updates.                                                                                                          \\ \hline
\textbf{SplitNN}                                                                & Low              & Low                & Low             & Synchronous client-server updates.                                                                                                          \\ \hline
\end{tabular}
\vspace{0.8em}
\caption{In this table, we compare the resources required for computation, bandwidth, memory and synchronization by techniques like federated learning, large batch synchronous SGD and SplitNN. SplitNN consumes fewer resources than federated learning and large batch synchronous SGD in these aspects except for synchronization requirements that are similar across all three techniques.}\label{resource_table2}
\end{table}
\vspace{-4em}
\subsection{Future Trends} Data poisoning attacks on federated learning \cite{fung2018mitigating} where malicious users can inject false training data to negatively effect the classification performance of the model have been proposed. Adversarial robustness to such attacks need to be improved. Using neural-network compression schemes in conjunction with federated learning to reduce the communication overload is an avenue for future work. Looking at combinations of federated learning and differential privacy, secure multi-party computation is an interesting direction for future work given that there has been active research in the recent time in all these areas. 

\section{Large Batch Synchronous SGD}

\subsection{Key Idea} The technique introduces additional backup workers to work on updating the weights, and chooses to synchronously update the aggregated model, as soon as any of the fastest N workers finish their updates. This is an improvement in accuracy over asynchronous SGD where some of the local workers might be updating the weights of a more stale model as the client-server updates are asynchronous. It also is relatively faster than synchronous SGD with no back-up workers where the servers wait for all the clients to finish their updates before aggregating the model parameters to update the model.

\subsection{Benefits} It allows for faster synchronous SGD, and is more accurate than asynchronous SGD approaches where some clients end up updating the weights based on a more stale model.

\subsection{Limitations} The computational requirements and communication bandwidth required is much higher than other distributed deep learning methods.

\subsection{Future Trends} The future trends are similar to that of federated learning as this method is very similar in essence to federated learning although it instead runs on a single batch of data. This method has high computational overload and network footprint. To make this method more sustainable in data center or decentralized settings, future work in improving its efficiency is important.

\begin{algorithm}[H]
\caption{Large-Batch SGD}
\begin{algorithmic}
\SUB{Worker Update($k$), where $k=1,\dots,N+b$}
\STATE \textbf{Input}: Dataset $\mathcal{X}$, $B$ mini-batch size.
\FOR{$t = 0,1,\dots$}
  \STATE Wait to read $\btheta^{(t)} = (\btheta^{(t)}[0], \dots, \btheta^{(t)}[M])$ $\quad\quad$from parameter servers.
  \STATE $G_k^{(t)} := 0$
  \FOR{$i = 1,\dots,B$}
    \STATE Sample datapoint $\widetilde{x}_{k,i}$ from $\mathcal{X}$.\;
   \STATE $G_k^{(t)} \leftarrow G_k^{(t)} + \frac{1}{B} \nabla F(\widetilde{x}_{k,i}; \btheta^{(t)})$.
  \ENDFOR
  Send $(G_k^{(t)}, t)$ to parameter servers.
\ENDFOR
\SUB{Parameter Server Update($j$), where $k=1,\dots,N+b$}
\STATE \textbf{Input} $\gamma_0,\gamma_1,\dots$ learning rates, $\alpha$ decay rate, $N$ number of mini-batches to aggregate, $\btheta^{(0)}$ model initialization.
\FOR{$t = 0, 1, \dots$}
  \STATE $\mathcal{G} = \{\}$\;
  \WHILE{$|\mathcal{G}| < N$}
    \STATE Wait for $(G, t')$ from any worker.\;
    \IF{$t' == t$}
        \STATE $\mathcal{G} \leftarrow \mathcal{G} \cup \{G\}$.
    \ELSE
    \STATE Drop gradient $G$.
    \ENDIF
  \ENDWHILE
  \STATE $\btheta^{(t+1)}[j] \leftarrow \btheta^{(t)}[j] - \frac{\gamma_t}{N} \sum_{G\in\mathcal{G}} G[j]$.\;
  \STATE $\bar\btheta^{(t)}[j] = \alpha \bar\btheta^{(t-1)}[j] + (1 - \alpha) \btheta^{(t)}[j]$.
\ENDFOR
\end{algorithmic}
\end{algorithm}

\section{Split Learning (SplitNN)}

\subsection{Key Idea} In this method each client trains the network upto a certain layer known as the cut layer and sends the weights to server. The server then trains the network for rest of the layers. This completes the forward propagation. Server then generates the gradients for the final layer and back-propagates the error until the cut layer. The gradient is then passed over to the client. The rest of the back-propagation is completed by client. This is continued till the network is trained. The shape of the cut could be arbitrary and not necessarily, vertical. In this framework as well there is no explicit sharing of raw 
data.

\subsection{Benefits}

Client-side communication costs are significantly reduced as the data to be transmitted is restricted to first few layers of the splitNN prior to the split. The client-side computation costs of learning the weights of the network are also significantly reduced for the same reason. In terms of model performance, the accuracies of Split NN remained much higher than federated learning and large batch synchronous SGD with a drastically smaller client side computational burden when training on a larger number of clients.
 
\subsection{Limitations} It requires a relatively larger overall communication bandwidth when training over a smaller number of clients although it ends up being much lower than other methods in settings with large number of clients. Advanced neural network compression methods such as \cite{lin2017deep,han2015deep,louizos2017bayesian} can be used to reduce the communication load. The communication bandwidth can also be traded for computation on client by allowing for more layers at client to represent further compressed representations.
\vspace{-1.7em}
\\\subsection{Future Trends} Given its no peek properties, no model detail sharing and high resource efficiency of this recently proposed method, it is well placed to provide direct applications to important domains like distributed healthcare, distributed clinical trials, inter and intra organizational collaboration and finance. Using neural-network compression schemes in conjunction with splitNN to reduce communication overload is a promising avenue for future work. Looking at combinations of federated learning and differential privacy, secure multi-party computation is an interesting direction for future work as well given that there has been active research in recent time in all these areas.

\begin{algorithm}[H]
\begin{algorithmic}
\SUB{Server executes at round $t\geq 0$:}
     \FOR{each client $k \in S_t$ \textbf{in parallel}}
       \STATE $\bA^k_{t} \leftarrow \text{ClientUpdate}(k, t)$ 
       \STATE Compute $\bW_{t} \leftarrow \bW_{t} - \eta \grad \loss(\bW_{t}; \bA_{t})$
       \STATE Send $\grad \loss(\bA_{t}; \bW_{t})$ to client $k$ for $\text{ClientBackprop}(k, t)$
     \ENDFOR
   \STATE

 \SUB{ClientUpdate($k, t$):}\ \ \  // \emph{Run on client $k$}
 \STATE $\bA^k_{t}= \phi$
  \FOR{each local epoch $i$ from $1$ to $\lepochs$}
    \FOR{batch $b \in \mathcal{B}$}
      \STATE Concatenate $f(b, \bH^k_t)$ to $\bA^k_{t}$
    \ENDFOR
 \ENDFOR
 \STATE return $\bA^k_t$ to server
\STATE
 \SUB{ClientBackprop($k, t, \grad \loss(\bA_{t}; \bW_{t})$):}\ \ \  // \emph{Run on client $k$}
\FOR{batch $b \in \mathcal{B}$}
  \STATE $\bH^k_t =  \bH^k_t - \eta \grad \loss(\bA_{t}; \bW_{t}; b)$
\ENDFOR
 
\end{algorithmic}
\mycaptionof{algorithm}{\algfont{SplitNN}\xspace. The $\nc$
  clients are indexed by $k$; $\lbs$ is the local minibatch size, and $\eta$ is the learning
  rate.}\label{alg:fedavg}
\end{algorithm}
\vspace{-1em}\section{Methods to Further Reduce Leakage and Improve Efficiency}

\subsection{Obfuscation with Differential Privacy for NN}

\subsubsection{Key Idea:} The methods in \cite{geyer2017differentially} modify stochastic gradient descent (SGD) based optimization used in learning neural networks by clipping the gradient for each lot of data and adding Gaussian noise to it during the optimization as opposed to adding noise to final parameters of the model, which could be overly conservative thereby affecting the utility of the trained model.The sigma for the noise is chosen at each step so as to maintain a guaranteed epsilon-delta differential privacy for a given lot of data. The tradeoff between the conflicting objectives of accuracy and privacy is determined by the lot size.      
\subsubsection{Benefits and Limitations:} The privacy is always dependent on a limited privacy budget while this also has an inversely proportional dependency with model accuracy. This is unlike in SplitNN where high accuracies are achieved without sharing raw data. The guarantees of differential privacy are currently theoretically backed unlike in SplitNN or Federated Learning. It also violates the no-peak rule when the privacy budget is over.
\vspace{-1em}
\subsubsection{Future Trends:} There is a lot of scope in combining differential privacy with distributed deep learning methods like splitNN, federated learning and large batch SGD as it adds to stronger theoretical guarantees on preventing data leakage.

\subsection{Homomorphic Encryption for NN}

\subsubsection{Key Idea} Homomorphic encryption aims to preserve the structure of ciphers such that addition and multiplicative operations can be performed after the encryption. All operations in a neural network except for activation functions are sum and product operations which can be encoded using Homomorphic encryption. Activation functions are approximated with either higher degree polynomials, Taylor series, standard or modified Chebyshev polynomials that are then implemented as part of Homomorphic encryption schemes. The works in \cite{hesamifard2017cryptodl,aono2018privacy} apply these ideas in the context of deep learning. A greatly detailed survey comparing various software libraries for homomorphic encryption is provided in \cite{vepakomma2018HESurvey}.

\newcommand{\calL}{\ensuremath{\mathcal L}}
\newcommand{\g}{\ensuremath{\mathbf g}}
\newcommand{\Id}{\ensuremath{\mathbf I}}
\begin{algorithm}[H]
	\caption{Differentially private SGD}\label{alg:privsgd}
	\begin{algorithmic}
	\REQUIRE Examples $\{x_1,...,x_N\}$, loss function $\calL(\btheta)=\frac{1}{N}\sum_i \calL(\btheta, x_i)$. Parameters: learning rate $\eta_t$, noise scale $\sigma$, group size $L$, gradient norm bound $C$. 
		\STATE {\bf Initialize} $\btheta_0$ randomly
		\FOR{$t \in [T]$}
		\STATE {Take a random sample $L_t$ with sampling probability $L/N$}
		\STATE {\bf Compute gradient}
		\STATE {For each $i\in L_t$, compute $\g_t(x_i) \leftarrow \nabla_{\btheta_t} \calL(\btheta_t, x_i)$}		
		\STATE {\bf Clip gradient}
		\STATE {$\bar{\g}_t(x_i) \gets \g_t(x_i) / \max\big(1, \frac{\|\g_t(x_i)\|_2}{C}\big)$}
		\STATE {\bf Add noise}
		\STATE {$\tilde{\g}_t \gets \frac{1}{L}\left( \sum_i \bar{\g}_t(x_i) + \mathcal{N}(0, \sigma^2 C^2 \Id)\right)$}
		\STATE {\bf Descent}
		\STATE { $\btheta_{t+1} \gets \btheta_{t} - \eta_t \tilde{\g}_t$}
		\ENDFOR
		\STATE {\bf Output} $\btheta_T$ and compute the overall privacy cost $(\varepsilon, \delta)$ using a privacy accounting method.
	\end{algorithmic}
\end{algorithm}\vspace{-2em}

\subsubsection{Algorithm}

There are a variety of schemes which have been shown to have homomorphic properties, and are provably secure. The most common use the security of the LWE (learning with errors) problem, which seeks to solve a linear system after adding noise. This problem is difficult to solve in certain conditions (when the dimension of the vector space is much larger than the computational range), and has even been shown to be secure under known quantum attacks. In short, LWE contains an algebraic structure with homomorphisms for addition and multiplication under the finite field of integers (so all multiplication/addition of finite integers can be encrypted and evaluated homomorphically as a LWE problem). In practice, implementations use R-LWE (Ring-LWE, use polynomial rings instead of vector spaces explicitly), which uses a slightly different representation, but the underlying algebraic structure remains largely the same.\\\textbf{Simple LWE example:} The key generation, encryption/decryption and corresponding add/multiply operations for a simple LWE example are given below.

\textbf{Keygen:}
     \begin{align}
     A \in \mathbb{Z} ^{m\times n}_{q}\nonumber\\ S\sim \mathbb{Z} ^{n}_{q}\nonumber\\  e\sim \mathcal{N} ^{n}\nonumber\\ b=As+e
     \end{align}
\\\textbf{Encrypt/Encode:}
\begin{align}
r_{1},e_{1} \sim \mathcal{N}\nonumber \\ c = \left( c_{a}, c_{b} \right) = \left( A^{T}r_{1}, b^{T}r_{1}+m_{1}+e_{1}\right) 
\end{align}
\\\textbf{Add/Multiply:}
\begin{align}c_{\text{add}}=c_{1}+c_{2}\\ c_{\text{mult}}= D \left( c_{1}\otimes c_{2} \right) \end{align} where $\otimes$ is the tensor product and $D$ is a dimension switching matrix that simplifies the resulting ciphertext. A proof of correctness and further sophistication addressing this scheme as a practical system can be found in the LWE literature.

\subsubsection{Benefits and Limitations}

These techniques need specialized hardware or extensive computational resources to scale. They are capable of providing a higher level of security that allows for perfect decryption and are not dependent on trade-offs of obfuscation vs. accuracy. The tradeoffs involved in this case are more with regards to computational efficiency. For example, some work (Microsoft’s SEAL) shows that to perform logistic regression on 1MB of data, 10GB of memory are required, and massive parallelization is necessary to achieve real-time throughput on practical problems (some tasks may not be parallelizable as such). LWE hardness is believed to be valid even in a post-quantum cryptographic environment.
                       
\subsubsection{Future Trends} This method requires very high computational resources, to make it scalable to practical deep learning architectures. Current techniques have only been benchmarked on simple networks over small datasets such as MNIST hand-written digit recognition. Development of faster methods for large scale deep learning and specialized hardware is an important avenue for future work. 

\subsection{Multi-Party Computation (MPC) and Garbled Circuits}
 
\subsubsection{Key Idea:}

These techniques are based on the idea of secret sharing and zero knowledge proof that we describe. The protection is achieved by sharing a secret message with different entities and requiring that these entities cooperate together in order to gain accessibility. There are certain problems where two entities collaborate to compute a function without sharing information about the inputs to the function with each other. The classic example is the millionaire’s problem, where $f(x_1,x_2)$ is computed by two parties, when one party has $x_1$ and the other has $x_2$, and it’s impossible for the party to the learn the value the other party holds. $f(x_1,x_2)$ will return a positive number if $x_1 > x_2$ and a negative value if $x_2 > x_1$. In this way, two millionaire’s can determine who has more money without sharing the total value at each hold. This has practical applications to untrusted “credit checks” or as an example for a particular kind of “Zero Knowledge Proof.” Yao’s \cite{YaoProtocols} garbled circuit protocol for the millionaire’s problem and 1–2 oblivious transfer \cite{rabin2005exchange,even1985randomized} are prominent works in this direction. Computational implementations and frameworks for this work such as Obliv-C, ObliVM, SPDZ and Sharemind are prominent.
\vspace{-1em}
\subsubsection{Benefits and Limitations:} These techniques have been studied for problems like secure stable matching, linear system solving and parallel graph algorithms. There is not much work done at intersection of MPC with deep learning.
\vspace{-1em}
\subsubsection{Future Trends:} Specialized hardwares for MPC are being developed to realize practical applications of these protocols. As current day machine learning relies heavily on large scale deep-learning architectures on large datasets, bridging this gap between MPC frameworks and distributed deep learning is an important avenue for future work.

\begin{table}[htbp]
\centering
\begin{tabular}{|l|l|l|}
\hline
\textbf{Method}    & \textbf{100 Clients} & \textbf{500 Clients} \\ \hline
\textbf{Large Batch SGD}    & 29.4 TFlops          & 5.89 TFlops          \\ \hline
\textbf{Federated Learning} & 29.4 TFlops          & 5.89 TFlops          \\ \hline
\textbf{SplitNN}            & 0.1548 TFlops        & 0.03 TFlops          \\ \hline
\end{tabular}
\vspace{1em}
\caption{Computation resources consumed per client when training CIFAR 10 over VGG (in teraflops) are drastically lower for SplitNN than Large Batch SGD and Federated Learning.}
\end{table}
\begin{table}[!htbp]
\centering
\begin{tabular}{|l|l|l|}
\hline
\textbf{Method}    & \textbf{100 Clients} & \textbf{500 Clients} \\ \hline
\textbf{Large Batch SGD}    & 13 GB                & 14 GB                \\ \hline
\textbf{Federated Learning} & 3 GB                 & 2.4 GB               \\ \hline
\textbf{SplitNN}            & 6 GB                 & 1.2 GB               \\ \hline
\end{tabular}

\caption{Computation bandwidth required per client when training CIFAR 100 over ResNet (in gigabytes) is lower for splitNN than large batch SGD and federated learning with a large number of clients. For setups with a smaller number of clients, federated learning requires a lower bandwidth than splitNN. Large batch SGD methods popular in data centers use a heavy bandwidth in both settings.}
\end{table}\vspace{-2.2em}
\section{Comparison of resource efficiency across no peek distributed deep learning}
We now share a comparison from \cite{gupta2018distributed} of validation accuracy and required client computational resources in Figure 1 for the three techniques of federated learning, large batch synchronous SGD and splitNN as they are tailored for distributed deep learning. 
\vspace{-0.2em}
As seen in this figure, the comparisons were done on datasets of CIFAR 10 and CIFAR 100 using VGG and Resnet-50 architectures for 100 and 500 client based setups respectively. In this distributed learning experiment we clearly see that SplitNN outperforms the techniques of federated learning and  \vspace{-1em}large batch synchronous SGD in terms of higher accuracies with drastically lower computational requirements on the side of clients. In tables 4 and 5 we share more comparisons from \cite{gupta2018distributed} on computing resources in TFlops and communication bandwidth in GB required by these techniques. SplitNN again has a drastic improvement of computational resource efficiency on the client side. In the case with a relatively smaller number of clients the communication bandwidth required by federated learning is less than splitNN.\vspace{-1em}
\begin{figure}[!htbp]%
    \centering
   \subfloat[Accuracy vs client-side flops on 100 clients with VGG on CIFAR 10]{ \includegraphics[width=5.5cm,height=6cm]{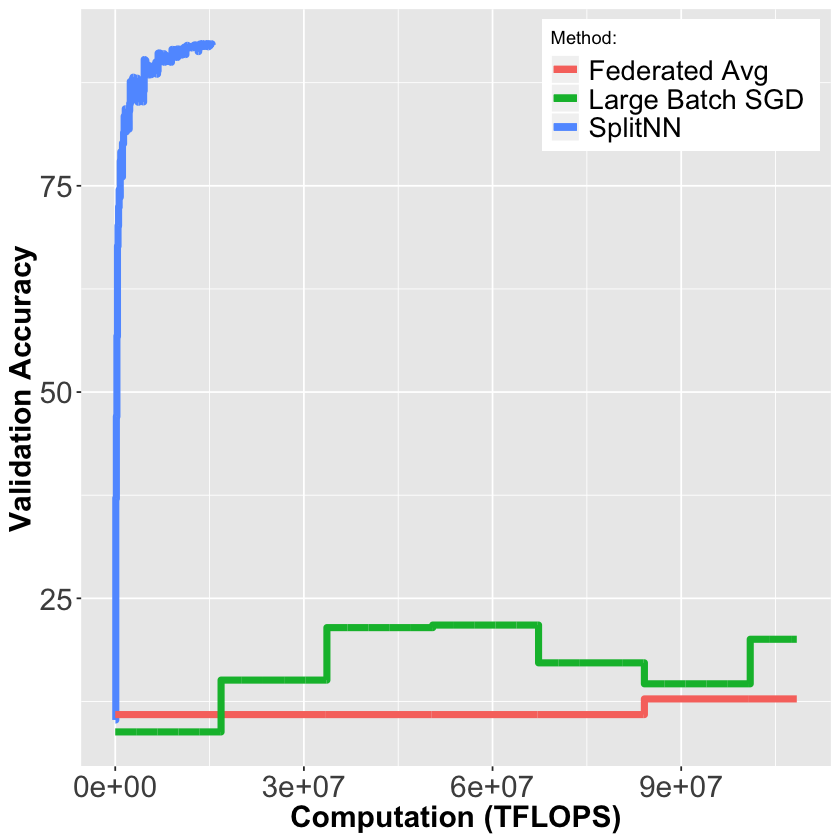}}
   \qquad
   \subfloat[Accuracy vs client-side flops on 500 clients with Resnet-50 on CIFAR 100]{ \includegraphics[width=5.5cm,height=6cm]{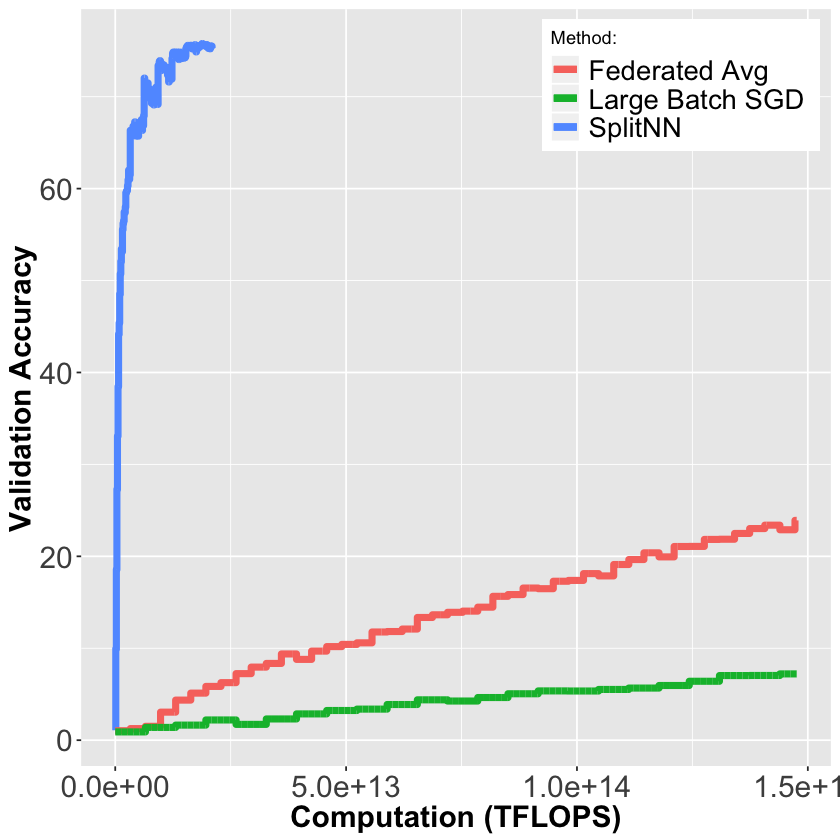}}%
     \caption{We show dramatic reduction in computational burden (in tflops) while maintaining higher accuracies when training over large number of clients with splitNN. Blue line denotes distributed deep learning using splitNN, red line indicate federated averaging and green line indicates large batch SGD. 
}%
\end{figure}
 \section{Conclusion and Future Work}
 No peek deep neural networks require new thinking when compared to existing data protection methods that attempt to aggregate siloed data for the benefit of server models. We describe the emergence of three methods in this setting: splitNN, federated learning and large batch SGD. Novel combinations of these methods with differential privacy, homomorphic encryption and secure MPC could further exploit theoretical guarantees. We show that in settings with large number of clients, splitNN needs the least communications bandwidth while federated learning does better with relatively smaller number of clients. In this direction, improving resource and communication efficiencies of no peek methods would be another avenue for impactful future work. Using advanced neural network compression methods \cite{lin2017deep,louizos2017bayesian,han2015deep} will help further reduce the required network footprint. It is also important to study adversarial robustness to data poisoning attacks \cite{fung2018mitigating} where malicious users can inject false training data to negatively effect the classification performance of the model. Adversarial attack schemes from this parallel research area need to be taken into consideration while developing no peek mechanisms. Efficient no peek methods have direct applications to important domains like distributed healthcare, distributed clinical trials, inter and intra organizational collaboration and finance. We therefore contemplate novel no peek distributed deep learning applications in the future.

  \end{document}